# Computer Vision Estimation of Emotion Reaction Intensity in the Wild

Yang Qian
University of Hawai'i at Manoa

Ali Kargarandehkordi
University of Hawai'i at Manoa

Onur Cezmi Mutlu
Stanford University

Saimourya Surabhi
Stanford University

Mohammadmahdi Honarmand
Stanford University

Dennis Paul Wall*
Stanford University

Peter Washington*
University of Hawai'i at Manoa

*Corresponding authors*

## Abstract

*Emotions play an essential role in human communication. Developing computer vision models for automatic recognition of expression can aid in a variety of domains, including robotics, digital behavioral healthcare, and media analytics. There are three types of emotional representations that are traditionally modeled in affective computing research: Action Units, Valence Arousal (VA), and Categorical Emotions. As part of an effort to move beyond these representations towards more fine-grained labels, we describe our submission to the newly introduced Emotional Reaction Intensity (ERI) Estimation challenge in the 5th competition for Affective Behavior Analysis inthe-Wild (ABAW). We developed four deep neural networks trained in the visual domain to predict emotion reaction intensity. Our best-performing model on the Hume-Reaction dataset achieved an average Pearson correlation coefficient of 0.31 on the test set using a pre-trained ResNet50 model. This work provides a first step towards the development of production-grade models which predict emotion reaction intensities rather than discrete emotion categories.*

## 1. Introduction

Natural facial expressions are the most potent, universally recognized signals for conveying emotional states and intentions [1, 2]. Mental disease diagnosis, human social/physiological interaction detection, social robotics, and many other sociotechnical applications have been target domains for research on automatic emotion recognition [3–8].

An example of affective computing which is central to our research groups' collective work is digital therapeutics for developmental delays. Emotional expressions play a crucial role in certain types of pediatric developmental disorders. Autism spectrum disorder (ASD), for example, affects almost 1 in 44 people in the United States [9], and it is one of the fastest-growing developmental disorders in terms of prevalence [10,11]. Children with autism tend to evoke emotions differently than neurotypical peers, and it is more difficult for them to produce the correct facial expressions [12–14]. Digital therapeutics have been developed to assist children who struggle with emotion by providing realtime emotion cues corresponding to the evocations of a conversational partner using real-time computer vision recognition of emotion expression [15–23]. Such digital and wearable devices enable families to provide therapy in the comfort of their homes and ability to customize the intervention structure to suit their child's needs [24–30]. However, these systems use models which only predict basic emotion categories and could benefit from the development of more sophisticated emotion recognition models.

Hume-React is a large-scale multimodal database containing user-generated video content and corresponding annotations of emotion reaction intensity.

By releasing the Hume-Reaction dataset, Hume contributed to the 5th Workshop and Competition on Affective Behavior Analysis InThe-Wild (ABAW) and to affective computing research as a whole. The released dataset includes more than 75 hours of video recordings consisting of spontaneous reactions of 2,222 individuals to 1841 evocative video elicitors. Each video is annotated by individuals with seven self-reported emotions at a scale of intensity 1-100 [31]. Prediction of continuous intensity rather than category alone can expand the possibilities of digital therapeutics for ASD.

We propose an affect recognition and level estimation model for the Emotional Reaction Intensity (ERI) Estimation task in the 5th ABAW Competition [32]. In contrast to the most recent ABAW competitions, where multi-task learning was the central theme or among one of the main challenges [33, 34], this year the focus is on only uni-task solutions to four challenges: Valence-Arousal (VA) Estimation [35, 36], Expression (Expr) Classification [37, 38], Action Unit (AU) Detection [37, 39], and Emotional Reaction Intensity (ERI) Estimation. We designed our algorithms to surpass the baseline network performance ResNet50 (pre-trained VGGFACE2) with fixed convolutional weights and employed multiple modifications to enhance the proposed models and achieve better efficiency in detecting emotion labels and estimating their intensity levels. Our code is publicly available here: https://github.com/YangQiantwx/EERI_CVPR_2023.

## 2. Related Work

A multitude of features, including visual, audio, text, and physiological signals, have been introduced in prior multimodal deep learning models. We focus on reviewing visual features, as it was the modality used in this project.

We first describe common visual feature representations. The Facial Action Coding System (FACS) is a widely used affect recognition network that recognizes specific emotions based on facial Action Units (AU) [40]. Gabor wavelet is another emotion recognition tool successfully applied to facial representation [41]. Benefiting from the growing application of deep learning, researchers have discovered that extracting features based on deep learning techniques can achieve higher accuracy. For instance, to extract visual features, [42] uses CNN and RNN stack based on a convolutional recurrent neural network. To prove the efficiency of audio/visual networks, [43] proposes the usage of 2D+1D convolutional neural networks. In the AVECs, researchers use other deep learning methods that all perform better than traditional feature extractors [44–46].

There are several prior works for facial emotion recognition [4,7,8,47,48]. A major bottleneck in affective computing is that emotion expression models are limited by the datasets they are trained on. Existing emotion datasets use only one of the three common types of emotional representations: Categorical Emotions (CE), Action Units (AU), and Valence Arousal (VA). The similarities between some expressions (i.e., the ambiguity of the labels in the dataset) is another challenge that increases the difficulty of distinguishing some facial expressions. This ambiguity might originate from inconsistent labeling. For example, "Sadness" can be similar to "Disgust" and it might be difficult to distinguish these two facial expressions.

To tackle some of the mentioned limitations above, the use of video-based datasets with alternative labeling representations has emerged. Dealing with the complexities of high-dimensional video data becomes a central challenge. Due to rapid expression changes, many frames might not contain reliable information for predicting facial expressions let alone estimation of the emotion's intensity. Labeling the video frame by frame [49] adds further complexity.

The Aff-Wild [50–53] and Aff-Wild2 [32, 54–58] Audio/Visual (A/V) datasets are current examples used in both academic and industrial communities that contain all three representation labels mentioned above. Aff-Wild2 is comprised of 548 videos of around 2.7M frames annotated in terms of the seven primary expressions (i.e., anger, disgust, fear, happiness, sadness, surprise, and neutral).

The 5th Workshop and Competition on Affective Behavior Analysis in-the-wild (ABAW) introduces four primary challenges. These include: 1) Valence-Arousal (VA) Estimation i.e., how positive/negative and active/passive an emotional state is, 2) Expression (Expr) Classification, and 3) Action Unit (AU) Detection (specific movements of facial muscles from Facial Action Coding System), and 4) a new 4th challenge called Emotional Reaction Intensity (ERI) Estimation using a new dataset of emotion reaction intensities (Hume Reaction).

## 3. Methodology

We train deep neural networks which use convolutional feature extractors [59] pretrained on a small sample dataset from AffectNet [60] to represent visual features. We optimize all models using Mean Squared Error (MSE) loss:

$$MSE = \frac{1}{n}\sum_{i=1}^{n}(y_i - \hat{y_i})^2$$

To evaluate model performance, we measure the Pearson correlation coefficient between the predicted emotion intensity and the ground truth intensity label for each emotion and calculate their average scores:

$$\rho = \frac{\sum(x_i - \bar{x})(y_i - \bar{y})}{\sqrt{\sum(x_i - \bar{x})^2 \sum(y_i - \bar{y})^2}}$$

$$\bar{p} = \frac{1}{n}\sum_{i=1}^{n} p_i$$

### 3.1. Data Preprocessing And Normalization

RetinaFace [61] was used to detect faces in each frame. The detected face coordinates were then used to crop the input frames to only show the detected faces, helping to prevent overfitting irrelevant portions of the image.

32 evenly separated frames were sampled for each video. The resulting shape of our training data is a tensor with shape [x, 32, 112, 112, 3], where x is the total number of data points used for training. We applied three data augmentation techniques for face analysis and emotion recognition [62]: brightness increases of up to 150% of the maximum value, horizontal flipping and rotation (20% range).

The data were normalized by subtracting the mean ($\mu$) of each feature and dividing by the standard deviation ($\sigma$), ensuring that each feature has a mean of $\theta$ and a standard deviation of 1:

$$Z_i = \frac{x_i - \mu}{\sigma}$$

## 4. Model Architecture

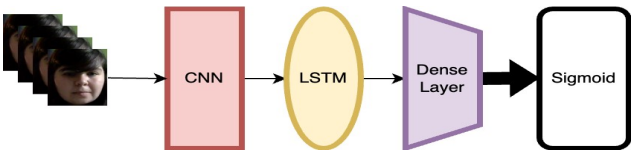

Figure 1. CNN-LSTM architecture.

We designed two types of frameworks for our network architecture. Both of them utilize convolutional neural networks for feature extraction. To interpret the features across timestamps, the first framework relies on long short-term memory (LSTM) network [63, 64] and the second one is based on Transformer [46].

### 4.1. CNN-LSTM

Our first proposed network architecture, termed the CNN-LSTM framework (Figure 1), is primarily characterized by its two key components - a Convolutional Neural Network (CNN) and a Long Short-Term Memory (LSTM) network. This pairing of feature extraction and temporal analysis forms an effective solution for processing spatially complex sequential data.

Initially, the input data passes through the CNN component of our model. This stage transforms the high dimensional input into a reduced feature space, generating a feature map that encapsulates the salient aspects of the input data. This feature map is then fed into the LSTM layer. The LSTM processes these features across time steps, capturing the temporal dependencies within the sequential data.

Following the LSTM stage, the extracted temporal features are further processed through a dense layer. This layer acts as a feature processor that distills the temporal features into a refined format. Lastly, another dense layer takes the refined features and transforms them into the final output. This layer employs a sigmoid activation function to predict the final emotion reaction intensities in the range of [0, 1].

### 4.2. CNN-Transformer

In the second framework, we incorporate a Convolutional Neural Network (CNN) with a Transformer model (Figure 2) to perform complex spatial and temporal analysis.

The framework begins with applying a CNN to the input tensor, which has the shape of ($T,H,W,3$) where $T$ denotes the number of frames, $H$ and $W$ are the height and width of each frame. The CNN extracts the spatial features from the input, resulting in a feature tensor of shape ($T,h,w,d$), where h and w are the reduced height and width, and d is the depth of the feature map.

To effectively capture spatial relationships in these features, we pass them through a spatial embedding layer, which reduces the dimensionality of the tensor to ($T,h,w,d'$). Following this, we perform global average pooling to remove the spatial dimensions, thus resulting in a feature vector of size ($T,d'$).

Next, we utilize position encoding to the feature vector and feed it into a TransformerEncoder layer characterized by parameters: $d_{ff}$ (size of the feed-forward layer in the Transformer block), $d_{model}$ (embedding dimension) = d', and num heads (number of attention heads). The output from this step maintains the shape (T, d') and can be expressed as:

$$X_{encoded} = Transformer(X_{pos}, d_{ff}, d_{model}, num_{heads})$$

where $X_{pos}$ is the position-encoded feature vector.

The output from the TransformerEncoder is then fed into a 1D Temporal Convolutional Network (TCN) for another layer of encoding, this time considering the temporal dependencies within the feature sequence. Following this, we again apply position encoding to the TCN output and pass it through another TransformerEncoder layer, using the same parameters as before.

Finally, we perform global average pooling on the output from the second TransformerEncoder layer. This pooled output is then passed through two consecutive dense layers. The final layer applies a sigmoid activation function to yield the final output.

## 5. Experiments

In our experiments, we trained and tested four distinct models - two using a pre-trained ResNet18 and two using a pre-trained ResNet50 for the CNN backbones.

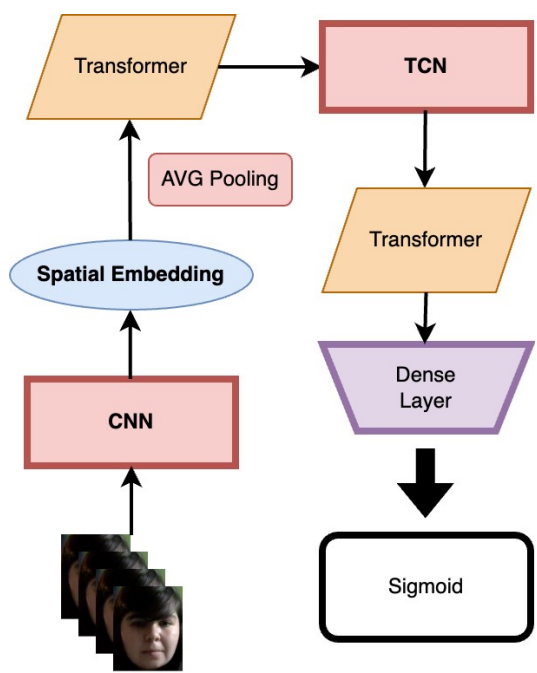

Figure 2. CNN-Transformer architecture.

### 5.1. Hardware and Software Setup

All computations were executed on an Nvidia A100 GPU, with models built using Keras. Our initial learning rate was set at 0.0002 and the total epoch is 50.

### 5.2. Model Configuration

The LSTM layer comprised 512 units. For the Transformer, we set the number of layers to 3, the embedding dimension ($d_{model}$) to 256, the size of the feed-forward layer ($d_{ff}$) to 128, and the number of attention heads to 8.

### 5.3. Training Process

The Adam optimizer was employed for training, with a batch size of 128. Training was monitored using three callback functions:

- EarlyStopping, which terminated training and restored the best model weights when the validation Pearson correlation failed to improve by a minimum delta of 0.0001 over 12 epochs.

- ReduceLROnPlateau, which reduced the learning rate by a factor of 0.5 when the validation Pearson correlation did not improve by at least 0.0001 over 6 epochs.

- ModelCheckpoint, which saved the model weights in a file whenever an improvement in the validation Pearson correlation was detected.

## 6. Results

The performance of each model on the Hume-Reaction dataset is provided in Table 1. It was evaluated based on the Pearson Correlation Coefficient. The baseline model achieved a Pearson Correlation Coefficient of 0.249.

### 6.1. Model Performance

Table 1 summarizes the performance of our models in terms of the Pearson Correlation Coefficient (PCC).

| Model | PCC |
|---|---|
| Baseline | 0.249 |
| ResNet18 + LSTM | 0.248 |
| ResNet50 + LSTM | 0.259 |
| ResNet18 + Transformer | 0.286 |
| ResNet50 + Transformer | 0.312 |

Table 1. Performance of models using Pearson Correlation Coefficient (PCC).

The ResNet50 + Transformer model outperformed all the other models and the baseline, achieving the highest Pearson Correlation Coefficient of 0.312. Meanwhile, the ResNet18 + LSTM model slightly underperformed when compared to the baseline. The results demonstrate the effectiveness of using Transformer models over LSTM when combined with CNNs, especially with ResNet50 as the base.

## 7. Conclusion

We evaluated a series of neural network architectures for emotional reaction intensity estimation using the Hume Reaction dataset. Among the architectures tested, our model leveraging the combination of a ResNet50 and Transformer (referred as R50 + Trans.) achieved the highest performance when operating on visual data solely. In contrast, the model coupling ResNet18 with LSTM (R18 + LSTM) displayed slightly inferior results compared to our baseline, implying the distinct advantage of Transformer over LSTM in this task.

The superior performance of the R50 + Trans. model can be attributed to the comprehensive integration of CNN and Transformer. This robust architecture is capable of capturing both spatial attributes from CNN and temporal dependencies from Transformer, thereby allowing effective estimation of emotional intensity.

In our future work, we plan to explore additional enhancements to our model. We will investigate alternative feature representations, refine model architectures, optimize hyperparameters, and experiment with different data fusion strategies. We expect that these efforts will further boost the model's effectiveness in predicting emotional reaction intensity. Despite the significant improvement brought by our best model, this is just the beginning of our journey towards a more sophisticated and accurate estimation of emotional intensity.

## Acknowledgements

The technical support and advanced computing resources from University of Hawaii Information Technology Services – Cyberinfrastructure, funded in part by the National Science Foundation CC* awards # 2201428 and # 2232862 are gratefully acknowledged.